\documentclass[letterpaper, 10 pt, conference]{ieeeconf}
\IEEEoverridecommandlockouts
\usepackage{times}
\usepackage[nomarginpar, margin=1in]{geometry}
\usepackage[utf8]{inputenc}
\usepackage{amsmath}
\newcommand{\probP}{\text{I\kern-0.15em P}}
\usepackage{amsfonts}
\usepackage{graphicx}
\usepackage{wrapfig}
\let\labelindent\relax
\usepackage{enumitem}
\usepackage[table,xcdraw]{xcolor}
\usepackage{multicol}
\usepackage{booktabs} 
\usepackage{multirow}
\usepackage[bookmarks=true]{hyperref}

\newcommand\smalldots{\ifmmode\ldots\else\makebox[1em][c]{.\hfil.\hfil.}\thinspace\fi}

\definecolor{Gray}{RGB}{240,240,240}
\newcolumntype{a}{>{\columncolor{Gray}}c}

\usepackage{xr}
\makeatletter
\newcommand*{\addFileDependency}[1]{
\typeout{(#1)}
\@addtofilelist{#1}
\IfFileExists{#1}{}{\typeout{No file #1.}}
}\makeatother

\newcommand*{\myexternaldocument}[1]{%
\externaldocument{#1}%
\addFileDependency{#1.tex}%
\addFileDependency{#1.aux}%
}
\myexternaldocument{supplementary}
\usepackage{xcolor}

\renewenvironment{thebibliography}[1]{
  \begin{oldthebibliography}{#1}
    \setlength{\itemsep}{0em}
    \setlength{\parskip}{0em}
}
{
  \end{oldthebibliography}
}

\title{\LARGE \bf
Learning Human Preferences Over Robot Behavior as Soft Planning Constraints
}

\author{Austin Narcomey$^{1*}$, Nathan Tsoi$^{1*}$, Ruta Desai$^{2}$ and Marynel V\'{a}zquez$^{1}$%
\thanks{This work was partially supported by Meta and the National Science Foundation (NSF) under Grant No. (IIS-2106690).}
\thanks{$^{1}$A. Narcomey, N. Tsoi and M. V\'azquez are with Yale University. *Equal contribution {\tt\small austin.narcomey@yale.edu}}%
\thanks{$^{2}$R. Desai is with Meta AI. {\tt\small rutadesai@meta.com}}%
}

\begin{document}

\maketitle
\thispagestyle{empty}
\pagestyle{empty}

\begin{abstract}
Preference learning has long been studied in Human-Robot Interaction (HRI) in order to adapt robot behavior to specific user needs and desires. Typically, human preferences are modeled as a scalar function; however, such a formulation confounds critical considerations on how the robot should behave for a given task, with desired -- but not required -- robot behavior. In this work, we distinguish between such required and desired robot behavior by leveraging a planning framework. Specifically, we propose a novel problem formulation for preference learning in HRI where various types of human preferences are encoded as soft planning constraints. Then, we explore a data-driven method to enable a robot to infer preferences by querying users, which we instantiate in rearrangement tasks in the Habitat 2.0 simulator. We show that the proposed approach is promising at inferring three types of preferences even under varying levels of noise in simulated user choices between potential robot behaviors. Our contributions open up doors to adaptable planning-based robot behavior in the future.
\end{abstract}

\section{Introduction}
\label{sec:introduction}

Robot learning methods have long been studied in Human-Robot Interaction (HRI) \cite{billard2008survey,chernova2014robot,thomaz2016computational,ravichandar2020recent} in order to adapt robot behavior to specific user needs and desires. A popular line of research pertains to encoding human preferences into a scalar function that represents \textit{hard constraints} which describe essential aspects for the success of the robot on the task of interest, and \textit{soft constraints} which describe how a human may like a robot to behave while executing the task. Soft constraints are important, but are not a requirement for the robot's success in completing the task.
For instance, if the function is a reward function, then higher output values signal better robot behavior in terms of solving the task or aligning with the user's preferences. Such a function can be used to evaluate the suitability of robot actions and optimize for robot behavior \cite{biyik2022aprel,wirth2017survey}.

\begin{figure}[t]
\centering
\small
\includegraphics[width=\linewidth]{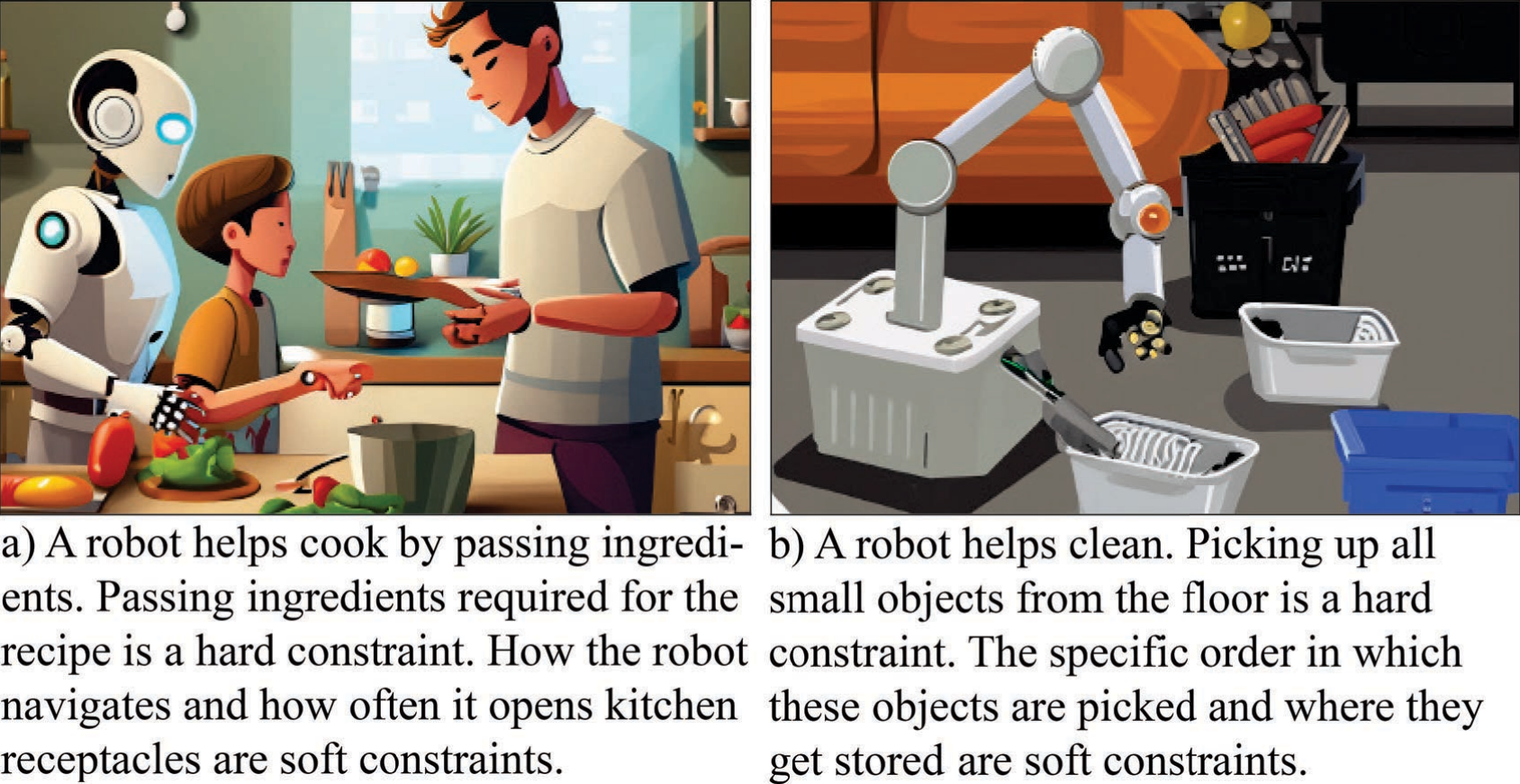}
\vspace{-15px}
\caption{
Illustrations of scenarios in which differentiating hard and soft task constraints can be beneficial for robot behavior generation. 
}
\vspace{-15px}
\label{fig:pull_scenarios}
\end{figure}

In this work, we make an explicit distinction between hard constraints and soft constraints because there are many human-robot interaction scenarios where hard constraints are often known to robot designers and fixed. Meanwhile, soft constraints are variable from user to user and even over time, as illustrated in Figure \ref{fig:pull_scenarios}.

More specifically, we leverage \textit{planning with preferences} \cite{jorge2008planning}   to generate robot behavior based on high-level actions (e.g., grab an object X, move an object X to a location Y, etc.). In this type of planning approach, hard constraints are described as planning goals. Soft constraints  are described in terms of properties of plans, or sequences of robot actions, that a user would prefer to be satisfied (e.g., in terms of action occurrences, temporal
relationship between the occurrence of actions). This  allows formally comparing if a plan is preferred over another. 

We focus on learning the soft constraints only and assume that the  hard task constraints are known to the robot.
In our experimental evaluation, preferences are represented as soft constraints in the Planning Domain Definition Language and considered when generating actions for a Fetch robot in the Habitat 2.0 simulator \cite{szot2021habitat}. We focus our work on rearrangement tasks because of their ubiquity across service robotics applications (e.g., robots in the home as in Fig.~\ref{fig:pull_scenarios}a and \ref{fig:pull_scenarios}b, in warehouses, etc.). Additionally, we focus on learning preferences using binary queries in which users indicate their preference among two trajectories of execution of robot behavior  because this interaction modality is popular in the HRI literature (e.g., \cite{Sadigh-RSS-17,jeon2020reward,fitzgerald2022inquire,hejna2023few}). Further, research suggests that this interaction modality  results in quality data and can be practical \cite{Sadigh-RSS-19,koppol2021interaction}. 

In summary, our contributions are: 1) a novel problem formulation for preference learning from human feedback in HRI based on planning-based robot behavior generation;  2) a data-driven method for estimating multi-objective  preferences as soft planning constraints; and 3) an evaluation of our approach under varied training and evaluation noise in Habitat 2.0 \cite{szot2021habitat}. Our work paves the path towards learning preferences as soft planning constraints  in real-world scenarios in the future.

\section{Related Work}
\label{sec:related_work}
\noindent
\textbf{Preference Learning.}
The problem of learning human preferences has long been studied, especially in recommendation engines \cite{furnkranz2010preference}. Typically, preferences are elicited via ranking or comparisons \cite{ailon2012active,biyik2019asking,wirth2017survey}, which motivated us to consider binary queries in our work.
While we incorporate binary user preferences as soft constraints in a planning-based approach, prior work has often incorporated preference choices in a Reinforcement Learning (RL) paradigm using various types of handcrafted reward functions~\cite{wilson2012bayesian, Sadigh-RSS-17, biyik2018batch, biyik2022aprel} or learned rewards~\cite{christiano2017deep}.
Various methods for preference learning with RL have been proposed \cite{lin2020review}, with a particularly popular method being Inverse Reinforcement Learning (IRL) \cite{ziebart2010modeling,sadigh2018planning}, where a reward function is estimated from expert demonstrations. IRL is often expensive for online robot adaptation because, internally, it needs to repeatedly solve RL problems.
Thus in this work, we explore learning preferences as soft constraints in planning. Our approach is similar in spirit to \cite{mantik2022preference} which aims to learn preferences for planning under the assumption of a linear mapping between observations and one type of preferences. However, we consider a more complex observation space, several types of preferences, and no assumptions regarding the mathematical relationship between the observations and preferences.

\vspace{0.5em}
\noindent
\textbf{Interactive Robot Learning.}
Robot behavior can be learned or adapted using different types of interaction modalities \cite{jeon2020reward}, which can be explicit or implicit.
Explicit interaction modalities include demonstrations of desired robot behavior~\cite{billard2008survey} and comparison queries \cite{Sadigh-RSS-17,jeon2020reward,fitzgerald2022inquire} whereas implicit communication occurs via social cues like gaze~\cite{saran2019enhancing,admoni2017social} or facial reactions~\cite{cui2021empathic,candon2024react}.
We consider explicit feedback provided through comparison queries because this interaction modality is popular in HRI  (e.g., \cite{Sadigh-RSS-17,jeon2020reward,fitzgerald2022inquire,hejna2023few}). Accurately answering comparison queries tends to be easier for non-expert robot users than providing high-quality demonstrations of robot behavior \cite{Sadigh-RSS-19}. Further, the queries tend to be less cognitively demanding and more preferred than providing corrections or labels for how a robot executes a task \cite{koppol2021interaction}. While we consider a setup in which a robot gathers information from a user at constant intervals, active learning setups where the robot decides how to query are interesting future work \cite{cakmak2010designing,taylor2021active,fitzgerald2022inquire}.

\vspace{0.5em}
\noindent
\textbf{Planning Robot Behavior.}
Our work builds on the Planning Domain Definition Language (PDDL) \cite{gerevini2005plan,pddl3.1} and is inspired by prior work on combining planning and learning-based methods in robotics (e.g., \cite{chiang2019rl,pokle2019deep,lee2020guided}) as well as other recent work on applying neural networks to program synthesis \cite{graves2014neural,parisotto2016neuro,chen2021latent}, including the synthesis of PDDL  \cite{simonnatural21,liu2023llm+}. 
While planning requires domain knowledge \cite{leonetti2016synthesis}, it is more easily interpretable than black-box policies such as those computed using state-of-the-art deep learning~\cite{christiano2017deep, hejna2023few,kim2023preference}.
In the future, interpretable planning approaches could facilitate explaining robot behavior to humans~\cite{dechant2023learning,ahn2022can,paxton2019prospection}.

\section{Proposed Approach}
\label{sec:approach}

\begin{figure*}[t!p]
    \centering
    \includegraphics[width=.85\linewidth]{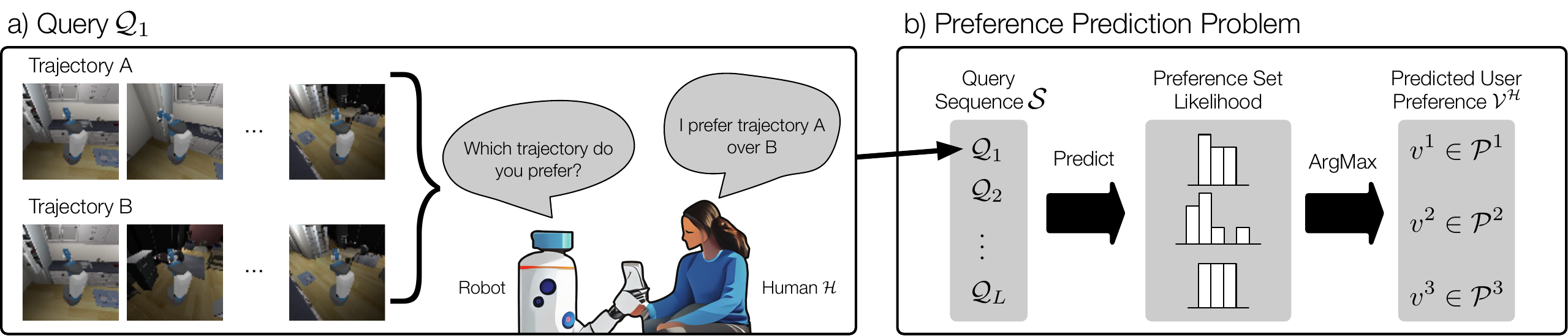}
    \vspace{-5px}
    \caption{Overview of interaction setup and approach. a) A query has trajectories A and B that show the robot completing the task of interest and a user's choice.
    b) Preference prediction problem. The robot's goal is to identify the user's preferences $\mathcal{V}^\mathcal{H}$ given a sequence of queries.
    }
    \label{fig:approach}
    \vspace{-15pt}
\end{figure*}

\subsection{Problem Setup}
\label{sec:problem_definition}

The goal of the robot is to learn a human's preferences over its behavior. In general, we assume that the robot has a PDDL domain and problem description that defines the task of interest, including the goal of the task, the objects that exist in the robot's workspace, predicates that describe the current state of the world, and the actions the robot can take (each defined by pre-condition and post-condition predicates). 
The predicates effectively map the continuous state of the world to a discrete, symbolic representation $\mathbf{x}$ for the state.

The robot's objective is to transform the initial task state $\mathbf{x}_{\text{ini}}$ to a goal state by executing discrete actions $a_1, ..., a_T$ that best align with a given user's preferences. Hard task constraints are encoded in the goal states via a \texttt{goal\_predicate}. A preference is a \textit{soft constraint} in planning: a  ``condition on the trajectory generated by a plan that the user would prefer to see satisfied, but is prepared to accept if it is not satisfied because of the cost involved, or because of conflicts with other constraints or goals'' \cite{gerevini2009deterministic}. Similar to hard constraints, preferences are expressed in PDDL predicates. 

The robot needs to learn the preferences by interacting with the user in order to generate behavior that the person likes, because the task objective for the robot is known but the user's preferences are not known a priori. We assume that human preferences or task objective do not change over time and are not affected by these interactions.

\vspace{0.5em}
\noindent
\textbf{Preference Space:} 
We assume that the space of user preferences $\mathcal{P}$ can be organized into a collection of disjoint preference sets, or \textit{sub-preferences} to distinguish them from the full collection $\mathcal{P}$. That is, $\mathcal{P} = \bigoplus_{n=1 \ldots N} \mathcal{P}^n$ where $n$ is an index over $N$ sub-preferences and $\oplus$ is the set concatenation operator. Each sub-preference has a clear semantic meaning. For example, consider a task to be completed in a kitchen setting, as shown in Fig.~\ref{fig:pull_scenarios}(a), but with a single user. 
One sub-preference could represent information about whether the user prefers the robot to place fruit on the table before fetching remaining dishware. Another sub-preference could represent information about the user's desired state for the receptacles in the kitchen (such as leaving the cabinets closed). In this setting, a preference would be defined by two values, one for the preference about the order of sub-task completion and the other corresponding to the preference for the state of receptacles. In this manner, each preference includes information about how the user would like the robot to perform the task in relation to multiple semantic factors.

\vspace{0.5em}
\noindent
\textbf{User Preference:}
For a user $\mathcal{H}$, we describe their preference $\mathcal{V}^\mathcal{H} \in \mathcal{P}$ as a collection of $N$ elements where each element is in a sub-preference space: 
\begin{equation}
    \small
    \mathcal{V}^\mathcal{H} = \, < v^n \,|\,  1 \leq n \leq N \wedge v^n \in \mathcal{P}^n >
    \label{eq:human_pref}
\end{equation}
with $v^n$ being an index drawn from $\mathcal{P}^n = \{1,\smalldots,|\mathcal{P}^n|\}$.
Each element $v^n$ tells us about how a user cares about a specific aspect of the world represented in the space of $\mathcal{P}^n$. 
For instance, a sub-preference space representing the desired state of receptacles in the kitchen could take $|\mathcal{P}^n|=3$ values: 1) the user prefers the receptacles to be closed, 2) prefers the receptacles to be open, or 3) has no preference regarding the receptacles' state.
Each user ($\mathcal{H}$) has one and only one sub-preference value for each sub-preference space. 
Importantly, each $v^n \in \mathcal{V}^\mathcal{H}$ in eq.~(\ref{eq:human_pref}) inherently defines a cost $\mathbf{c}^n \in [-1, 1]$ over transitions in a trajectory in PDDL. 
The cost \textbf{c} serves as soft constraints for planning with PDDL. 
\footnote{
It is possible for $\mathcal{V}^\mathcal{H}$ to include conflicting sub-preferences. Imagine that a robot completes a sorting task in a kitchen with all receptacles open and the user has preferences $\mathcal{V}^\mathcal{H} = <v^1, v^2>$ across $2$ sub-preferences.  The sub-preference $v^1$ is ``minimize operating the receptacles'' and $v^2$ is ``keep the receptacles closed''. Which preference is more important depends on the underlying costs associated with them. In this work, we do not optimize over these costs, but assume they are fixed such that all sub-preferences are equally important (e.g., satisfying a receptacle operations preference is as important as satisfying a receptacle state preference).
}

\vspace{0.5em}
\noindent
\textbf{Human Feedback:} We focus on learning soft constraints for PDDL planning in scenarios where a  robot repeatedly queries the user about their preferences over its behavior. 
For a query $\mathcal{Q}$, we show the user
two trajectories executed by the robot $\{\xi_a, \xi_b\}$ that visually describe how it could solve the task.
That is, given the sequence of states ($\mathbf{x}$) and actions ($a$): $\xi_a = < \mathbf{x}_1, a_1, \ldots, \mathbf{x}>$ where $
\mathbf{x}_1 = \mathbf{x}_{\text{ini}}$ and the final state $\mathbf{x}$ satisfies \texttt{is\_true(goal\_predicate($\mathbf{x}$))}.
Similar to prior work on preference learning \cite{biyik2022aprel}, the user responds by choosing the preferred trajectory (or no preference), as shown in Fig.~\ref{fig:approach}(a). Hereafter, we refer to the data included in a query as a triplet $\mathcal{Q} = <\xi_a, \xi_b, c>$ where $c$ is the user's choice: either the first trajectory is preferred, $\xi_a \succ \xi_a$; the second one is preferred,  $\xi_a \prec \xi_b$; or the user has no preference for either, $\xi_a \sim \xi_b$. 

\subsection{Predicting Preferences}
\label{ssec:pred_pref}

We take advantage of modern simulation capabilities~\cite{szot2021habitat} in order to create a model that computes the likelihood of user preferences given a sequence of user queries. This model is a neural network, whose ultimate goal is to predict the specific preference $\mathcal{V}^\mathcal{H}$, as in eq.~(\ref{eq:human_pref}), of a given user who answers the sequence of queries  $\mathcal{S} = < \mathcal{Q}_1, \ldots, \mathcal{Q}_L >$.
We focus on learning the preference as efficiently as possible irrespective of the specific queries posed to the user or their order. 

At first glance, one may think of training the neural network using a target that specifies the one ground truth preference $\mathcal{V}^\mathcal{H} \in \mathcal{P}$.
However, doing so ignores the structure of the preferences, where $\mathcal{V}^\mathcal{H} \in \bigoplus_{n=1 \ldots N} \mathcal{P}^n$  and each value $v^n$ is in the sub-preference $\mathcal{P}^n$. Thus, we propose to implement the neural network with $N$ heads (outputs), each of which tries to predict the correct element $v^n$ as a classification problem within $\mathcal{P}^n$, as shown in Fig.~\ref{fig:approach}(b).

\begin{figure*}[t!p]
\centering
    \includegraphics[width=0.8\linewidth]{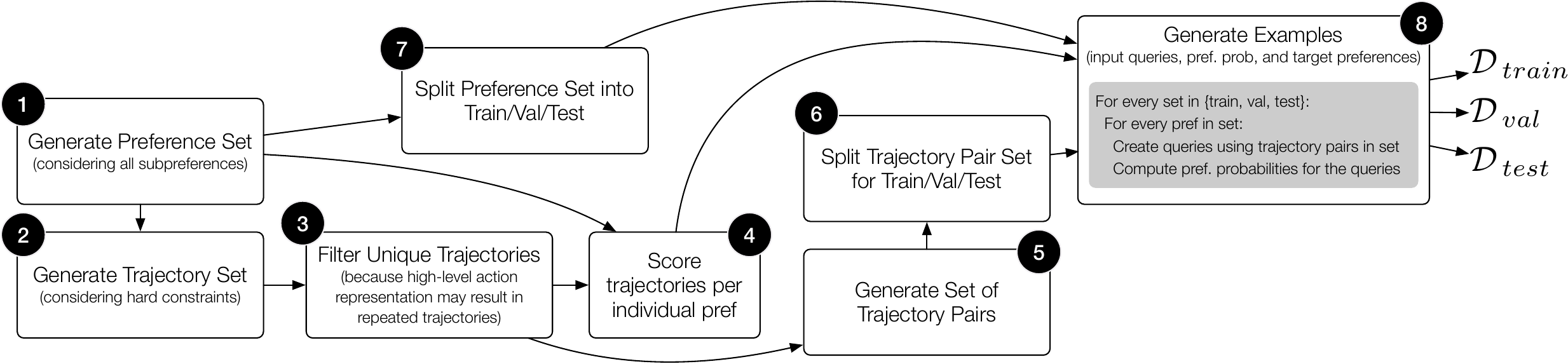}
    \vspace{-5pt}
    \caption{The proposed dataset processing pipeline used to generate our disjoint train, val, and test datasets. See Sec. \ref{ssec:supervision} and \ref{ssec:preference-prediction} for details.}
    \vspace{-15pt}
\label{fig:data-pipeline}
\end{figure*}

We hypothesize that a second challenge would arise when implementing the neural network. Because multiple preferences $\mathcal{V} \in \mathcal{P}$ may induce the same choice $c$ as $\mathcal{V}^\mathcal{H}$ for a given query $\mathcal{Q}$, the neural network could have trouble associating a query sequence $\mathcal{S}$ to the targets  $v^n \in \mathcal{V}^\mathcal{H}$. For example, consider a robot that has to complete a sorting task in a kitchen with a fridge and the sequence has a single query, $| \mathcal{S} | = 1$. The initial state for the task specifies that there are open receptacles. The user prefers that the robot leave receptacles closed, because it looks tidier. 
The user also prefers that the robot serves snacks by placing them on the table prior to setting dishware on the table. To collect the user's preferences regarding task execution order, the user is presented with two robot trajectories. In one trajectory, the robot closes receptacles when it is near that receptacle and serves snacks before setting dishware on the table. In the other trajectory, the robot leaves receptacles open and sets dishware on the table first before placing snacks on the table. The user chooses that they prefer the first trajectory over the second one. It is unclear from this choice whether the user prefers receptacles being closed, or prefers the robot to place snacks on the table before dishware, or both. Note that such ambiguity is unavoidable -- in order to ensure that user preference can be inferred from a reasonable number of queries, multiple preferences need to be expressed in each query. 
To address this challenge, we propose to train the neural network to predict probability distributions capable of capturing this uncertainty. The distributions describe the compatibility of human feedback with each sub-preference.

\subsection{Model Supervision}
\label{ssec:supervision}

The process that we use to create a dataset for training the proposed preference prediction model is presented in Fig.~\ref{fig:data-pipeline}. Let this dataset $\mathcal{D}$ contain $M$ examples, such that each example is composed of a query sequence $\mathcal{S}_m = < \mathcal{Q}_1, \ldots, \mathcal{Q}_L >_m$ with $L$ queries and a ground-truth preference $\mathcal{V}_m = <v^1, \smalldots, v^N >_m$:
\footnote{We make the assumption that the choices in the queries in $\mathcal{S}$ are consistent with the preference $\mathcal{V}_m$. For example, if $\mathcal{V}_m$ assigns a higher cost for trajectory $\xi_a$ than an alternative  $\xi_b$ for a given query, then the user's choice $c$ indicates that the user prefers $\xi_b$ over $\xi_a$. This assumption is later relaxed in our evaluation (Sec. \ref{sec:evaluation})}
\begin{equation}
 \small
 \mathcal{D}  
 = \big\{ 
 ( \mathcal{S}_1, \mathcal{V}_1), \smalldots, ( \mathcal{S}_M, \mathcal{V}_M)
 \big\}
 \label{eq:dataset}
\end{equation}
\noindent

We train neural network models using the features $\mathcal{S}$ and the ground-truth targets values $\mathcal{V}$.
We investigate supervising the training process using two types of data based on the target values.
First, we predict the target values directly, where each $v^n$ corresponds to a network head, using cross-entropy loss as described in Section \ref{ssec:pred_pref}.
Alternatively, we train models to predict a probability distribution  $\probP(v^n \,|\, \mathcal{S})$, which reflects the likelihood that $v^n \in \mathcal{P}^n$ is consistent with user all choices in the queries of $\mathcal{S}$.

To compute the probability $\probP(v^n \,|\, \mathcal{S})$, we identify all other examples in the dataset $\mathcal{D}$ which have different preferences $\mathcal{V}$ but have indistinguishable queries $\mathcal{S}$, as  $\probP(v^n  \,|\, \mathcal{S})$ should capture all of these indistinguishable preferences given a query sequence $\mathcal{S}$.

More formally, we compute the probability mass function (PMF) of $\probP(v^n|\mathcal{S})$ where $V^n$ is a discrete random variable with the range $\{ 1, \smalldots, |\mathcal{P}^n|\}$.
Let $\mathcal{C}_m$ be a set of examples consistent with example $m$, that is,
$\mathcal{C}_m = \{(\mathcal{S}_i, \mathcal{V}_i) \in \mathcal{D}|\mathcal{S}_i = \mathcal{S}_m\}$ for $i \in \{1, \smalldots M\}$, and let $y_i^n$ be a one-hot encoded vector of preferences, that is $y_i^n = < e_j \in \{0,1\}^{k^n}|j=(\mathcal{V}_i)^n>$ where $k \in \{1, \smalldots, |\mathcal{P}^n|\}$.
Then the PMF of $\probP(v^n|\mathcal{S})$ is defined by the events $\{V^n = k | k \in \{1, \smalldots, |\mathcal{P}^n|\}\} = \sum_{i=1}^{|\mathcal{C}_m|} \frac{1}{|\mathcal{C}_m|} y^{n}_{i,k}$ which occur with probability proportional to the number of instances of the consistent sub-preferences.
This PMF is used to supervise training via Kullback–Leibler (KL) divergence loss.

\section{Evaluation Setup}
\label{sec:evaluation}

\noindent
Our evaluation focuses on investigating three fundamental questions in a systematic manner using simulated human responses to robot queries:
\begin{description}[style=unboxed,leftmargin=0cm,parsep=0.5em]

\item [Q1.] \textit{How do noisy-rational user choices in robot queries affect preference prediction?} Prior work in economics and preference learning has shown that humans are not perfectly-rational decision makers. In particular, Hebert A. Simon \cite{simon1955behavioral} proposed that humans make choices based on the information they have available to them and their computational capacity, rather than simply maximizing utility based on their preferences. Such a \textit{bounded rationality} model for human decision-making has been widely used in the preference learning literature to simulate human choices \cite{lee2021b,biyik2022aprel, jeon2020reward}. Thus, we use it in our work to investigate how noisy-rational choices affect preference prediction.

\item[Q2.] \textit{How well do models that are trained on one level of noise in query choices generalize to data with a different noise level?} People can deviate from optimal decision making in many ways \cite{chan2021human}, making it hard to predict how noisy feedback, reflected by the user's choices in $\mathcal{S}$, may affect model performance on an unseen query sequence.
Thus, we investigate how models trained on varied levels of noise in the query choices perform when evaluated on data with a different level of noise.

\item[Q3.] \textit{How is preference prediction performance affected by the model's supervision?}
Since there may be multiple preferences \{$\mathcal{V}_1, \mathcal{V}_2, \smalldots$\} in $\mathcal{D}$ that are consistent with a sequence of user choices $\mathcal{S}$, we compare the performance of preference prediction models either supervising on the distributions of preferences $\probP(v^n \,|\, \mathcal{S})$, where $v^n \in \mathcal{P}^n$, or supervising on single ground truth targets $<v^1,\smalldots,v^N>$, as described in Sec. \ref{ssec:supervision}.

\end{description}

\subsection{Experimental Setup}
\label{sec:preference-summary}

\noindent
\textbf{Preferences in Habitat: }
We used the Habitat 2.0 simulator \cite{szot2021habitat} to generate robot behavior and simulate user choices between pairs of robot trajectories. In particular, we used the ``set\_table" rearrangement task in our work, where the robot can navigate an apartment environment from the Replica CAD dataset~\cite{szot2021habitat} as well as place objects in their designated target locations.

In our experiments, we consider $N=3$ sub-preferences in a rearrangement task in a kitchen environment: 1) the final state of the kitchen receptacles, 2) the number of receptacles  that the robot operated, and 3) the order of the robot completing sub-tasks. 
To create a diverse dataset considering these sub-preferences, we adapted Habitat-Lab~\cite{szot2021habitat}, a modular high-level library for end-to-end AI development with Habitat 2.0, such that we could plan with constraints and preferences in PDDL3.0 and PDDL3.1~\cite{pddl3.1, kovacs2012multi} using the OPTIC planner \cite{benton2012temporal}. This involved creating a new PDDL domain definition for high-level planning to support an interface to generate plans with OPTIC as a robot tries to solve a rearrangement task in Habitat 2.0. Further, we extended this new PDDL domain with additional actions and predicates for the sub-preferences mentioned earlier. 

\vspace{0.5em}
\noindent
\textbf{Simulating Noisy User Choices:} 
\label{ssec:noisy_choices}
We simulate human choices in binary queries using a bounded rationality model \cite{simon1955behavioral, jeon2020reward}. The likelihood of a human $\mathcal{H}$ 
selecting a query trajectory increases exponentially as the costs of the trajectory decreases: $P(\xi) \propto e^{-\beta \text{cost}(\xi)}$. Here, the lower the cost, the more the trajectory aligns with the human's preferences $\mathcal{V}^\mathcal{H}$ per eq. (\ref{eq:human_pref}). More specifically, the $\text{cost}(\xi)$for a trajectory is the sum of costs $\mathbf{c}^n$ for all sub-preferences $v^n \in \mathcal{V}^\mathcal{H}$.

In our experiments, we considered $\beta=\{10,1,0.5\}$, where $\beta=1$ is a typical noise level~\cite{barnett2023active}, $\beta=10$ resulted in less noise, and $\beta=0.5$ led to more noise in query choices. We also consider the case of no noise (identified as $\beta=\emptyset$ later in our results).
We introduce the noise in our data by randomly substituting choices within each $\mathcal{S}$ in $\mathcal{D}$, creating $\mathcal{S}_{\beta=\{10, 1, 0.5\}}$.

There are two ways we construct probability distributions used for model supervision. In both cases, the construction of the distribution depends on the consistency of query sequences $\mathcal{S}$, that is: $\mathcal{C}_m = \{(\mathcal{S}_i, \mathcal{V}_i) \in \mathcal{D}|\mathcal{S}_i = \mathcal{S}_m\}$.
First, non-noisy distributions are constructed from $\mathcal{S}$, which does not contain noise, as in Section \ref{ssec:supervision}.
Second, noisy distributions are constructed from $\mathcal{S}_{\beta=\{10, 1, 0.5\}}$, which have added noise.

\vspace{0.5em}
\noindent
\textbf{Evaluation Measures:} We consider two measures to evaluate preference prediction results: \textit{True Positive Rate} (TPR) and \textit{Accepted Output Rate} (AOR). 
TPR is the fraction of correctly predicted preferences. A preference ($\mathcal{V}$) is correct only when all sub-preference values $<~v^1, \smalldots, v^N~>$ are also correctly predicted.
AOR is a more permissive measure and is equal to the fraction of predicted preferences that are consistent with the corresponding non-noisy queries provided to a prediction model (consistency as defined in Sec.~\ref{ssec:supervision}). AOR indicates how ``reasonably" well a model predicts preferences consistent with the queries.

\subsection{Preference Prediction Models}
\label{ssec:preference-prediction}

We consider three types of models to predict human preferences based on an input sequence of queries $\mathcal{S}$:
\vspace{0.5em}
\begin{description}[style=unboxed,leftmargin=0cm,itemsep=0.5em]
\item[Perfectly Rational Baseline:]
The Perfectly Rational model draws a random preference from all preferences $\mathcal{V}$ in $\mathcal{D}$ that are consistent with all queries in $\mathcal{S}$, where consistency is computed as in Section \ref{ssec:supervision}.

\item[Neural Network Supervised on Single-Target Data:] 
We use a neural network to predict the ground-truth targets for each sub-preference $<v^1,\smalldots, v^N>$. We first use a bi-directional Long Short-Term Memory (LSTM) network~\cite{schmidhuber1997long} to embed each query trajectory $\xi$ into a fixed size representation. The corresponding choices are embedded via lookup table and then concatenated with their corresponding trajectory representations from the LSTM. A multi-layered perceptron (MLP) with $N$ heads is then used to predict the sub-preferences in parallel.
We use stochastic mini-batch gradient descent to minimize the cross-entropy loss between the ground-truth target $v^n$ for each sub-preference space and the network's $N$ heads. 
Let $\hat{y}^n$ be the network output from a head corresponding to sub-preference $n$, and let $\alpha^n$ be a coefficient weighting between sub-preferences. For a mini-batch of $B$ examples drawn from $D$:
\begin{equation}
    \small
    \ell_{ce} = - \sum_{i=1}^B \sum_{n=1}^{N} \alpha^n \, v^n_i \, \log \left( \hat{y}^n_i \right)
    \label{eq:cross-entropy}
\end{equation}

\item[Neural Networks Supervised on Distribution Data:] We train two models with the same neural network architecture as the single-target model. Both models are trained with a KL-Divergence loss, where the supervision minimizes the statistical distance between $\probP(v^n  \,|\, \mathcal{S})$ with $v^n \in \mathcal{P}^n$, and the network's softmax output for each sub-preference $\mathcal{P}^n$. But 
one model is trained using the non-noisy distribution constructed from $\mathcal{S}$ as in Sec. \ref{ssec:supervision}, while the other model uses a noisy distribution ($\mathcal{S}_{beta=10}, \mathcal{S}_{beta=1}, \text{or } \mathcal{S}_{beta=0.5}$ from Sec. \ref{ssec:noisy_choices}) as the ground-truth distribution. 
Specifically, consider an example $i$ in the training data. Let the ground truth probability mass for the example $i$ be $p_{V^n}(v^n_i)$, $\alpha^n$ the weighting coefficient between sub-preferences, and the network's predicted probability corresponding to the sub-preference value $v^n$ be $\hat{y}^n_i$. For a network with N heads, a mini-batch of B examples drawn from D:
\begin{equation}
    \small
    \ell_{kl} = -  \sum_{i=1}^{B} \sum_{n=1}^{N} \alpha_n \underbrace{  p_{V^n}(v^n_i) \, \log\left( \frac{p_{V^n}(v^n_i)}{\hat{y}^n_i} \right)}_{\text{loss for sub-preference $n$}}
    \label{eq:kl}
\end{equation}

\end{description}

\subsection{Neural Network Training Details}
\label{ssec:nn_training}
All neural networks were implemented in PyTorch and supervised with a multi-task objective where each component of the loss corresponds to a sub-preference.
The weight coefficients $\alpha$ were learned during training. 
We split dataset $\mathcal{D}$, of size $M=2,627,664$, into three disjoint datasets: $\mathcal{D}_{train}$, $\mathcal{D}_{val}$, $\mathcal{D}_{test}$ in the ratio 60/20/20.
Trajectory pairs {$\xi_a, \xi_b$} and user preferences $\mathcal{V}$ were not repeated across disjoint datasets. The data creation pipeline is summarized in Fig. \ref{fig:data-pipeline}.

Neural networks were trained on an NVIDIA 2080TI or a 3090TI GPU using the AdamW optimizer~\cite{loshchilov2017decoupled}. We used fast early stopping~\cite{prechelt2002early} at 60 successive steps of non-decreasing validation set loss, where each step is evaluated at 5\% intervals of an epoch on a random 5\% sample of validation data. 
We performed grid-search for batch size and learning rate hyper-parameters, and  report results on the complete test set.

\section{Results}
\label{sec:results}

\begin{table*}[tb!]
  \centering
  {\small
    \caption{True Positive Rate (TPR) and Accepted Output Rate (AOR) for 11 models trained using two types of supervision (distribution or single-target) and for the Perfect Rational (PR) baseline. PR does not involve learning. The  features input to the models correspond to 10 queries, which include choices over trajectory pairs. The amount of noise in the  choices of the evaluation and training data are denoted by $\beta_{eval}$ and $\beta_{train}$, respectively. The target distributions used for training were constructed using the choices without noise (\dag) or with noise (\ddag).}
    \vspace{-5pt}
    \label{tbl:combined-10q}
    {\footnotesize
    \begin{tabular}{lllccccccccccac}
    \multicolumn{2}{r}{\textit{Supervision:}} & \multicolumn{4}{c}{Single Target\dag} & \multicolumn{4}{c}{Distribution\dag} & \multicolumn{3}{c}{Noisy Distribution\ddag} & \multicolumn{1}{c}{PR} & \\
    \cmidrule(lr){3-6} \cmidrule(lr){7-10} \cmidrule(lr){11-13} \cmidrule(lr){14-14}
    \multicolumn{2}{r}{$\beta_{train}$:} & $\emptyset$ & $10$ & $1$ & $0.5$ & $\emptyset$ & $10$ & $1$ & $0.5$ & $10$ & $1$ & $0.5$ & \multicolumn{1}{c}{N/A} & \textit{row} \\ \midrule

    \parbox[t]{1mm}{\multirow{4}{*}{\rotatebox[origin=c]{90}{TPR}}} & $\beta_{eval}=\emptyset$ & 0.68 & 0.69 & 0.17 & 0.10 & 0.72 & 0.68 & 0.17 & 0.09 & 0.61 & 0.26 & 0.27 & 0.90 & (1) \\
    & $\beta_{eval}=10$ & 0.59 & 0.65 & 0.17 & 0.11 & 0.63 & 0.65 &  0.17 & 0.09 & 0.53 & 0.25 & 0.26 & 0.53 & (2) \\
    & $\beta_{eval}=1$ & 0.16 & 0.21 & 0.14 & 0.11 & 0.17 & 0.22 & 0.13 & 0.09 & 0.14 & 0.13 & 0.13 & 0.07 & (3)\\
    & $\beta_{eval}=0.5$ & 0.12 & 0.17 & 0.12 & 0.11 & 0.13 & 0.17 & 0.12 & 0.09 & 0.11 & 0.10 & 0.10 & 0.06 & (4) \\
    \midrule

    \parbox[t]{1mm}{\multirow{4}{*}{\rotatebox[origin=c]{90}{AOR}}} & $\beta_{eval}=\emptyset$ & 0.77 & 0.78 & 0.20 & 0.12 & 0.82 & 0.77 & 0.20 & 0.10 & 0.70 & 0.32 & 0.34 & 1.00 & (5) \\
    & $\beta_{eval}=10$ & 0.66 & 0.72 & 0.20 & 0.12 & 0.70 & 0.72 & 0.20 & 0.10 & 0.60 & 0.31 & 0.32 & 0.59 & (6) \\
    & $\beta_{eval}=1$ & 0.18 & 0.23 & 0.16 & 0.12 & 0.18 & 0.24 & 0.15 & 0.10 & 0.16 & 0.16 & 0.16 & 0.08 & (7) \\
    & $\beta_{eval}=0.5$ & 0.14 & 0.18 & 0.14 & 0.12 & 0.14 & 0.19 & 0.13 & 0.10 & 0.12 & 0.12 & 0.13 & 0.07 & (8) \\
    \end{tabular}
    }
    \vspace{-16pt}
  }
\end{table*}

Table \ref{tbl:combined-10q} reports the performance of 12 models, one per column, each trained using a sequence $\mathcal{S}$ with $L=10$ queries.
Models were supervised on targets without noise (non-noisy distribution or ground-truth target) and compared to the Perfectly Rational (PR) baseline described in Section \ref{ssec:preference-prediction}.
We also supervised models on the noisy distribution as the optimization target.
Each model was evaluated on held-out data with the same four noise levels considered for training. 

\textbf{RQ1: Effect of noisy-rational user choices}.
Given that humans are not perfectly-rational (PR) decision makers, we compare the performance of each model when trained and evaluated on data with the same noise level ($\beta_{eval}=\beta_{train}$). 
Our proposed method under-performs the PR baseline when trained and evaluated on no noise: (row $\beta_{eval}=\emptyset$, column $\beta_{train}=\emptyset$) $<$ (row $\beta_{eval}=\emptyset$, column PR ) in Table \ref{tbl:combined-10q})
Models trained on any amount of noise performed similarly or better than the PR baseline: for $\beta \in \{10,1,0.5\}$, (row $\beta_{eval}=\beta$, column $\beta_{train}=\beta$) $>=$ (row $\beta_{eval}=\beta$, column PR).

\textbf{RQ2: Generalization across different levels of noise.}
To evaluate the generalization capability of our approach, we evaluate the performance of each model on evaluation data with a different level of noise ($\beta_{train} \neq \beta_{eval}$).
Within types of supervision without noise (Distribution\dag, Single Target\dag), models trained on no noise performed better when evaluated on data without noise than when evaluated on data with noise:
(row $\beta_{eval}=\emptyset$, column $\beta_{train}=\emptyset$) $>$ (row $\beta_{eval}=\{10,1.0,0.5\}$, column $\beta_{train}=\emptyset$). However, models trained on a small amount of noise ($\beta_{train}=10$) performed better than models trained without noise when evaluated on data with noise: (row $\beta_{eval}=\{10,1.0,0.5\}$ column $\beta_{train}=10$) $>$ (row $\beta_{eval}=\{10,1.0,0.5\}$ column $\beta_{train}=\emptyset$). Further increasing the choice noise ($\beta_{train}=\{1,0.5\}$) in the training data did not improve performance.

\textbf{RQ3: Training Supervision.}
We first compare models supervised via distribution against models supervised on a single target (Distribution\dag~and Single Target\dag~in Table \ref{tbl:combined-10q}), both of which do not consider choice noise in the training targets.
When models were trained on data without noise and evaluated on all levels of noise including none, we found the prediction performance to be similar or better for models supervised via Distribution\dag~versus models supervised via Single Target\dag: (column $\beta_{train}= \emptyset$ Distribution\dag) $>=$ (column $\beta_{train}=\emptyset$ Single Target\dag).
Models trained on intermediate noise ($\beta_{train}=\{10, 1\}$) performed similarly regardless of supervision.
When training noise was high ($\beta_{train}=0.5$), models performed slightly better when supervised on a Single Target\dag~compared to Distribution\dag.

Finally, we compare models trained on the Noisy Distribution\ddag~with models trained on the non-noisy Distribution\dag.
With a low level of noise in the training data ($\beta_{train}=10$), Noisy Distribution\ddag~performed worse than non-noisy Distribution\dag~for all levels of evaluation noise.
However, when greater noise levels were present in the training data, $\beta_{train}=\{1,0.5\}$, then supervision via Noisy Distribution\ddag~outperformed supervision via non-noisy Distribution\dag: (column $\beta_{train}=\{1,0.5\}$ Distribution\ddag) $>$ (column $\beta_{train}=\{1,0.5\}$ Distribution\dag)

\section{Discussion}
\label{sec:discussion}

Our results show promise towards identifying preferences that could be applied as soft constraints in planning problems. 
We primarily considered models supervised on data that does not consider choice noise in the targets (Distribution\dag~or Single Target\dag~in Table \ref{tbl:combined-10q}). 
This revealed three insights.
First, regarding RQ1, noise in user choices resulted in decreased model performance. However, we found that models trained on at least some noise and evaluated on the same level of noise outperformed the PR baseline.
Regarding RQ2, we found the ability of a model to generalize to a different noise level depends on the noise level in the input features observed at training time.
Models trained without choice noise tended generalize poorly to data with noise.
Meanwhile, models trained on a small amount of choice noise ($\beta=10$) generalized better to other levels of noise.
Finally, regarding RQ3, we expected supervision via Distribution\dag~to outperform supervision via a Single Target\dag, which was often the case for models trained on with input features without noise. We were surprised to find that performance of Single Target\dag~vs. Distribution\dag~was similar when noise was present in the input features in the training data.

We considered training models using a Noisy Distribution\ddag~for the targets because real-world human feedback is likely to have some level of choice noise and computing non-noisy target distributions may not be possible in all applications due to the non-noisy choices being non-observable constructs in the real-world. While models supervised on Noisy Distribution\ddag~did not perform as well as models supervised on non-noisy Distribution\dag~when a small amount of choice noise was present in the training features, the results were different for more noise. When the amount of choice noise in the training data increased, models supervised on the Noisy Distribution\ddag~outperformed the models supervised on the non-noisy Distribution\dag.

\subsection{Limitations \& Future Work}

\noindent
Our work has limitations that point to interesting future directions. First, the sub-preferences considered in this work are a small subset of all possible preferences that can be expressed in PDDL.
Future work could expand the preference space and potentially explore learning new preferences in an online fashion.

Second, we predicted preferences using a set of pre-defined queries but, in the future, our preference prediction models could be used to generate more informative queries during online learning \cite{cakmak2010designing,fitzgerald2022inquire}. 
Also, one could leverage predicted distributions over sub-preferences to decide when to query a user, e.g., when there is high uncertainty about the user's preferences. 

Finally, we studied noisy-rational choices in our simulated experiments because they model real human feedback \cite{simon1990bounded}.
In the future, we would like to evaluate our approach in the real-world, identifying preferences from human feedback collected during collaborative human-robot tasks. This includes adapting robot behavior in-the-moment  \cite{brawer2023interactive} according to  preference estimates.

\section{Conclusion}

We investigated the problem of learning human preferences over robot behavior using preference-based planning in rearrangement tasks. The novelty of our work was three-fold: First, we framed preference learning by explicitly distinguishing between soft and hard constraints on robot behavior. This distinction was supported by the use of planning methods and the simulation capabilities of Habitat 2.0. Second, we investigated data-driven methods to predict preferences in a multi-objective supervised-learning paradigm, where the output of learned models corresponded to a sub-preference with specific semantic meaning.
Third, we studied the effects of training and evaluating on query data with noisy choices to the test generalization of our approach. 
Our results show that our approach is promising.
Learned models were able to infer simulated user preferences when there was noise in the features that described query data to varying degrees.
We are excited about continuing this line of work and testing our approach with real users in the future.

\bibliographystyle{IEEEtran}
%
\bibliography{references}

\begin{thebibliography}{10}
\providecommand{\url}[1]{#1}
\csname url@samestyle\endcsname
\providecommand{\newblock}{\relax}
\providecommand{\bibinfo}[2]{#2}
\providecommand{\BIBentrySTDinterwordspacing}{\spaceskip=0pt\relax}
\providecommand{\BIBentryALTinterwordstretchfactor}{4}
\providecommand{\BIBentryALTinterwordspacing}{\spaceskip=\fontdimen2\font plus
\BIBentryALTinterwordstretchfactor\fontdimen3\font minus \fontdimen4\font\relax}
\providecommand{\BIBforeignlanguage}[2]{{%
\expandafter\ifx\csname l@#1\endcsname\relax
\typeout{** WARNING: IEEEtran.bst: No hyphenation pattern has been}%
\typeout{** loaded for the language `#1'. Using the pattern for}%
\typeout{** the default language instead.}%
\else
\language=\csname l@#1\endcsname
\fi
#2}}
\providecommand{\BIBdecl}{\relax}
\BIBdecl

\bibitem{billard2008survey}
A.~Billard, S.~Calinon, R.~Dillmann, and S.~Schaal, ``Survey: Robot programming by demonstration,'' \emph{Springer handbook of robotics}, 2008.

\bibitem{chernova2014robot}
S.~Chernova and A.~L. Thomaz, \emph{Robot learning from human teachers}.\hskip 1em plus 0.5em minus 0.4em\relax Morgan \& Claypool Publishers, 2014.

\bibitem{thomaz2016computational}
A.~Thomaz, G.~Hoffman, M.~Cakmak \emph{et~al.}, ``Computational human-robot interaction,'' \emph{Found. Trends Mach. Learn.}, 2016.

\bibitem{ravichandar2020recent}
H.~Ravichandar, A.~S. Polydoros, S.~Chernova, and A.~Billard, ``Recent advances in robot learning from demonstration,'' \emph{Annu Rev Control Robot Auton Syst.}, 2020.

\bibitem{biyik2022aprel}
E.~B{\i}y{\i}k, A.~Talati, and D.~Sadigh, ``Aprel: A library for active preference-based reward learning algorithms,'' in \emph{HRI}, 2022.

\bibitem{wirth2017survey}
C.~Wirth, R.~Akrour, G.~Neumann, J.~F{\"u}rnkranz \emph{et~al.}, ``A survey of preference-based reinforcement learning methods,'' \emph{JMLR}, 2017.

\bibitem{jorge2008planning}
A.~Jorge, S.~A. McIlraith \emph{et~al.}, ``Planning with preferences,'' \emph{AI Magazine}, 2008.

\bibitem{szot2021habitat}
A.~Szot, A.~Clegg, E.~Undersander, E.~Wijmans, Y.~Zhao, J.~Turner, N.~Maestre, M.~Mukadam, D.~Chaplot, O.~Maksymets, A.~Gokaslan, V.~Vondrus, S.~Dharur, F.~Meier, W.~Galuba, A.~Chang, Z.~Kira, V.~Koltun, J.~Malik, M.~Savva, and D.~Batra, ``Habitat 2.0: Training home assistants to rearrange their habitat,'' in \emph{NeurIPS}, 2021.

\bibitem{Sadigh-RSS-17}
D.~Sadigh, A.~Dragan, S.~Sastry, and S.~Seshia, ``Active preference-based learning of reward functions,'' in \emph{RSS}, 2017.

\bibitem{jeon2020reward}
H.~J. Jeon, S.~Milli, and A.~Dragan, ``Reward-rational (implicit) choice: A unifying formalism for reward learning,'' \emph{NeurIPS}, 2020.

\bibitem{fitzgerald2022inquire}
T.~Fitzgerald, P.~Koppol, P.~Callaghan, R.~Q. J.~H. Wong, R.~Simmons, O.~Kroemer, and H.~Admoni, ``Inquire: Interactive querying for user-aware informative reasoning,'' in \emph{CoRL}, 2022.

\bibitem{hejna2023few}
D.~J. Hejna~III and D.~Sadigh, ``Few-shot preference learning for human-in-the-loop rl,'' in \emph{CoRL}, 2023.

\bibitem{Sadigh-RSS-19}
M.~Palan, G.~Shevchuk, N.~C. Landolfi, and D.~Sadigh, ``Learning reward functions by integrating human demonstrations and preferences,'' in \emph{RSS}, 2019.

\bibitem{koppol2021interaction}
P.~Koppol, H.~Admoni, and R.~G. Simmons, ``Interaction considerations in learning from humans.'' in \emph{IJCAI}, 2021, pp. 283--291.

\bibitem{furnkranz2010preference}
J.~F{\"u}rnkranz and E.~H{\"u}llermeier, ``Preference learning and ranking by pairwise comparison,'' in \emph{Preference learning}.\hskip 1em plus 0.5em minus 0.4em\relax Springer, 2010, pp. 65--82.

\bibitem{ailon2012active}
N.~Ailon, ``An active learning algorithm for ranking from pairwise preferences with an almost optimal query complexity.'' \emph{JMLR}, 2012.

\bibitem{biyik2019asking}
E.~B{\i}y{\i}k, M.~Palan, N.~C. Landolfi, D.~P. Losey, and D.~Sadigh, ``Asking easy questions: A user-friendly approach to active reward learning,'' \emph{arXiv:1910.04365}, 2019.

\bibitem{wilson2012bayesian}
A.~Wilson, A.~Fern, and P.~Tadepalli, ``A bayesian approach for policy learning from trajectory preference queries,'' \emph{NeurIPS}, 2012.

\bibitem{biyik2018batch}
E.~Biyik and D.~Sadigh, ``Batch active preference-based learning of reward functions,'' in \emph{CoRL}, 2018.

\bibitem{christiano2017deep}
P.~F. Christiano, J.~Leike, T.~Brown, M.~Martic, S.~Legg, and D.~Amodei, ``Deep reinforcement learning from human preferences,'' \emph{NeurIPS}, 2017.

\bibitem{lin2020review}
J.~Lin, Z.~Ma, R.~Gomez, K.~Nakamura, B.~He, and G.~Li, ``A review on interactive reinforcement learning from human social feedback,'' \emph{IEEE Access}, vol.~8, 2020.

\bibitem{ziebart2010modeling}
B.~D. Ziebart, J.~A. Bagnell, and A.~K. Dey, ``Modeling interaction via the principle of maximum causal entropy,'' in \emph{ICML}, 2010.

\bibitem{sadigh2018planning}
D.~Sadigh, N.~Landolfi, S.~S. Sastry, S.~A. Seshia, and A.~D. Dragan, ``Planning for cars that coordinate with people: leveraging effects on human actions for planning and active information gathering over human internal state,'' \emph{Auton. Robots}, 2018.

\bibitem{mantik2022preference}
S.~Mantik, M.~Li, and J.~Porteous, ``A preference elicitation framework for automated planning,'' \emph{Expert Systems with Applications}, vol. 208, p. 118014, 2022.

\bibitem{saran2019enhancing}
A.~Saran, E.~S. Short, A.~Thomaz, and S.~Niekum, ``Enhancing robot learning with human social cues,'' in \emph{HRI}, 2019.

\bibitem{admoni2017social}
H.~Admoni and B.~Scassellati, ``Social eye gaze in human-robot interaction: a review,'' \emph{JHRI}, 2017.

\bibitem{cui2021empathic}
Y.~Cui, Q.~Zhang, B.~Knox, A.~Allievi, P.~Stone, and S.~Niekum, ``The empathic framework for task learning from implicit human feedback,'' in \emph{Conf. on Robot Learning}.\hskip 1em plus 0.5em minus 0.4em\relax PMLR, 2021.

\bibitem{candon2024react}
K.~Candon, N.~C. Georgiou, H.~Zhou, S.~Richardson, Q.~Zhang, B.~Scassellati, and M.~V\'{a}zquez, ``React: Two datasets for analyzing both human reactions and evaluative feedback to robots over time,'' in \emph{HRI}, 2024.

\bibitem{cakmak2010designing}
M.~Cakmak, C.~Chao, and A.~L. Thomaz, ``Designing interactions for robot active learners,'' \emph{IEEE T Auton Ment De}, 2010.

\bibitem{taylor2021active}
A.~T. Taylor, T.~A. Berrueta, and T.~D. Murphey, ``Active learning in robotics: A review of control principles,'' \emph{Mechatronics}, vol.~77, 2021.

\bibitem{gerevini2005plan}
A.~Gerevini and D.~Long, ``Plan constraints and preferences in pddl3,'' Dept Elect Autom, U Brescia, Tech. Rep., 2005.

\bibitem{pddl3.1}
``{BNF definition of PDDL 3.1},'' \url{https://helios.hud.ac.uk/scommv/IPC-14/repository/kovacs-pddl-3.1-2011.pdf}, 2011.

\bibitem{chiang2019rl}
H.-T.~L. Chiang, J.~Hsu, M.~Fiser, L.~Tapia, and A.~Faust, ``Rl-rrt: Kinodynamic motion planning via learning reachability estimators from rl policies,'' \emph{RA-L}, 2019.

\bibitem{pokle2019deep}
A.~Pokle, R.~Mart{\'\i}n-Mart{\'\i}n, P.~Goebel, V.~Chow, H.~M. Ewald, J.~Yang, Z.~Wang, A.~Sadeghian, D.~Sadigh, S.~Savarese \emph{et~al.}, ``Deep local trajectory replanning and control for robot navigation,'' in \emph{ICRA}, 2019.

\bibitem{lee2020guided}
M.~A. Lee, C.~Florensa, J.~Tremblay, N.~Ratliff, A.~Garg, F.~Ramos, and D.~Fox, ``Guided uncertainty-aware policy optimization: Combining learning and model-based strategies for sample-efficient policy learning,'' in \emph{ICRA}, 2020.

\bibitem{graves2014neural}
A.~Graves, G.~Wayne, and I.~Danihelka, ``Neural turing machines,'' \emph{arXiv:1410.5401}, 2014.

\bibitem{parisotto2016neuro}
E.~Parisotto, A.-r. Mohamed, R.~Singh, L.~Li, D.~Zhou, and P.~Kohli, ``Neuro-symbolic program synthesis,'' \emph{arXiv:1611.01855}, 2016.

\bibitem{chen2021latent}
X.~Chen, D.~Song, and Y.~Tian, ``Latent execution for neural program synthesis beyond domain-specific languages,'' \emph{NeurIPS}, 2021.

\bibitem{simonnatural21}
N.~Simon and C.~Muise, ``A natural language model for generating pddl,'' in \emph{KEPS}, 2021.

\bibitem{liu2023llm+}
B.~Liu, Y.~Jiang, X.~Zhang, Q.~Liu, S.~Zhang, J.~Biswas, and P.~Stone, ``Llm+ p: Empowering large language models with optimal planning proficiency,'' \emph{arXiv:2304.11477}, 2023.

\bibitem{leonetti2016synthesis}
M.~Leonetti, L.~Iocchi, and P.~Stone, ``A synthesis of automated planning and reinforcement learning for efficient, robust decision-making,'' \emph{Artif Intell}, 2016.

\bibitem{kim2023preference}
C.~Kim, J.~Park, J.~Shin, H.~Lee, P.~Abbeel, and K.~Lee, ``Preference transformer: Modeling human preferences using transformers for {RL},'' in \emph{ICLR}, 2023.

\bibitem{dechant2023learning}
C.~DeChant, I.~Akinola, and D.~Bauer, ``Learning to summarize and answer questions about a virtual robot's past actions,'' \emph{arXiv:2306.09922}, 2023.

\bibitem{ahn2022can}
M.~Ahn, A.~Brohan, N.~Brown, Y.~Chebotar, O.~Cortes, B.~David, C.~Finn, C.~Fu, K.~Gopalakrishnan, K.~Hausman \emph{et~al.}, ``Do as i can, not as i say: Grounding language in robotic affordances,'' \emph{arXiv:2204.01691}, 2022.

\bibitem{paxton2019prospection}
C.~Paxton, Y.~Bisk, J.~Thomason, A.~Byravan, and D.~Foxl, ``Prospection: Interpretable plans from language by predicting the future,'' in \emph{ICRA}, 2019.

\bibitem{gerevini2009deterministic}
A.~E. Gerevini, P.~Haslum, D.~Long, A.~Saetti, and Y.~Dimopoulos, ``Deterministic planning in the fifth international planning competition: Pddl3 and experimental evaluation of the planners,'' \emph{Artif Intell}, 2009.

\bibitem{simon1955behavioral}
H.~A. Simon, ``A behavioral model of rational choice,'' \emph{QJE}, 1955.

\bibitem{lee2021b}
K.~Lee, L.~Smith, A.~Dragan, and P.~Abbeel, ``B-pref: Benchmarking preference-based reinforcement learning,'' \emph{arXiv:2111.03026}, 2021.

\bibitem{chan2021human}
L.~Chan, A.~Critch, and A.~Dragan, ``Human irrationality: both bad and good for reward inference,'' \emph{arXiv:2111.06956}, 2021.

\bibitem{kovacs2012multi}
D.~L. Kov{\'a}cs, ``A multi-agent extension of pddl3. 1,'' 2012.

\bibitem{benton2012temporal}
J.~Benton, A.~Coles, and A.~Coles, ``Temporal planning with preferences and time-dependent continuous costs,'' in \emph{ICAPS}, 2012.

\bibitem{barnett2023active}
P.~Barnett, R.~Freedman, J.~Svegliato, and S.~Russell, ``Active reward learning from multiple teachers,'' \emph{arXiv preprint arXiv:2303.00894}, 2023.

\bibitem{schmidhuber1997long}
J.~Schmidhuber, S.~Hochreiter \emph{et~al.}, ``Long short-term memory,'' \emph{Neural Comput}, 1997.

\bibitem{loshchilov2017decoupled}
I.~Loshchilov and F.~Hutter, ``Decoupled weight decay regularization,'' \emph{arXiv:1711.05101}, 2017.

\bibitem{prechelt2002early}
L.~Prechelt, ``Early stopping-but when?'' in \emph{Neural Networks: Tricks of the trade}.\hskip 1em plus 0.5em minus 0.4em\relax Springer, 2002, pp. 55--69.

\bibitem{simon1990bounded}
H.~A. Simon, ``Bounded rationality,'' \emph{Utility and probability}, pp. 15--18, 1990.

\bibitem{brawer2023interactive}
J.~Brawer, D.~Ghose, K.~Candon, M.~Qin, A.~Roncone, M.~V{\'a}zquez, and B.~Scassellati, ``Interactive policy shaping for human-robot collaboration with transparent matrix overlays,'' in \emph{HRI}, 2023.

\end{thebibliography}


\end{document}